\documentclass[runningheads]{llncs}

\usepackage{multirow}
\usepackage[ruled,linesnumbered]{algorithm2e}
\usepackage{amssymb}
\usepackage{amsmath}

\usepackage{booktabs, cellspace, tabularx}
\usepackage{multicol}
\usepackage{multirow}
\usepackage{pifont}
\usepackage{marvosym}
\usepackage{wrapfig}
\usepackage{float}                  
\usepackage{subfig}                 

\usepackage[table,xcdraw]{xcolor}

\usepackage[T1]{fontenc}
\usepackage[table,xcdraw]{xcolor}
\usepackage{graphicx}

\usepackage[pagebackref,breaklinks,colorlinks,citecolor=blue]{hyperref}

\begin{document}
\title{A New Perspective to Boost Performance Fairness For Medical Federated Learning
}
%
\author{Yunlu Yan$^{1}$, Lei Zhu$^{1,6}$ (\Letter), Yuexiang Li$^2$, Xinxing Xu$^3$, Rick Siow Mong Goh$^3$, 
Yong Liu$^3$, Salman Khan$^{4,5}$, Chun-Mei Feng$^3$}
\institute{$^1$ The Hong Kong University of Science and Technology (Guangzhou), \\ Guangzhou, China  \\
$^2$ Guangxi Medical University, Nanning, China \\
$^3$ Institute of High Performance Computing (IHPC),
Agency for Science, Technology and Research (A*STAR), Singapore\\
$^4$ Mohamed bin Zayed University of Artificial Intelligence (MBZUAI), UAE\\
$^5$ Australian National University, Canberra ACT, Australia\\
$^6$ The Hong Kong University of Science and Technology, Hongkong, China
\\ \email{leizhu@ust.hk} \\ \href{https://github.com/IAMJackYan/Fed-LWR}{\texttt{https://github.com/IAMJackYan/Fed-LWR}}}

\maketitle              
\begin{abstract}


Improving the fairness of federated learning (FL) benefits healthy and sustainable collaboration, especially for medical applications. However, existing fair FL methods ignore the specific characteristics of medical FL applications, \textit{i.e.}, domain shift among the datasets from different hospitals. In this work, we propose \texttt{Fed-LWR} to improve performance fairness from the perspective of feature shift, a key issue influencing the performance of medical FL systems caused by domain shift. Specifically, we dynamically perceive the bias of the global model across all hospitals by estimating the layer-wise difference in feature representations between local and global models. To minimize global divergence, we assign higher weights to hospitals with larger differences. The estimated client weights help us to re-aggregate the local models per layer to obtain a fairer global model. We evaluate our method on two widely used federated medical image segmentation benchmarks. The results demonstrate that our method achieves better and fairer performance compared with several state-of-the-art fair FL methods.



\end{abstract}

\section{Introduction}

Federated learning~\cite{li2020federated,mcmahan2017communication} (FL) has emerged as a hot research topic in healthcare~\cite{guan2023federated}, offering a framework for effectively leveraging diverse datasets to learn a better global model without compromising privacy, significantly facilitating communication and collaboration among medical institutions. However, most of the existing works~\cite{jiang2022harmofl,feng2022specificity,jiang2023iop,wang2023feddp,liu2021federated} primarily focus on improving the global performance (\textit{e.g.}, average accuracy), incurring disproportionately advantages or disadvantages on some institutions. Fairness has always been an important evaluation criterion for machine learning models, especially for FL applications. An unfair FL system can hurt the incentives of users to participate. 


Existing fairness-related FL studies primarily focus on two aspects: \textit{collaboration fairness}~\cite{lyu2020collaborative,xu2021gradient} and \textit{performance fairness}~\cite{li2019fair,li2021ditto}. The former pays more attention to resource allocation during the learning process, while the latter concerns the balance of model outcomes. Considering that most real-world users are consequentialists, they are more concerned about the intuitive performance of the learned model on their private dataset, rather than the intricate details of the learning process. Therefore, the main goal of this work is to improve the performance fairness of medical FL system. Performance fairness is often an overlooked concern when striving for higher global performance. For example, although the average accuracy may be high, the accuracy at some institutions could be significantly lower. This skew in performance outcomes necessitates a more nuanced approach to ensure equitable improvements across all participating entities in FL applications. However, overemphasizing fairness in performance is also nonsensical, as the goal of fair FL is not to achieve identical accuracy for every user. The key challenge lies in striking a trade-off between `high' and `fair'.

\begin{figure*}[t]
\centering
  \includegraphics[width=1.0\textwidth]{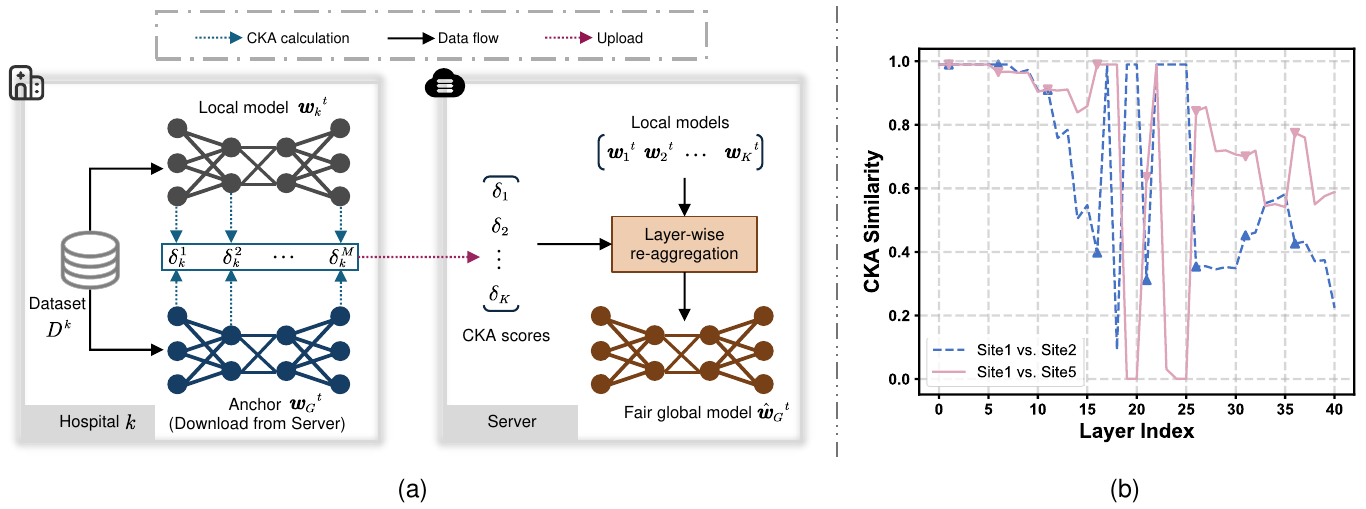} \caption{(a) \textbf{Overview of} \texttt{Fed-LWR}. During the parameter aggregation stage of $t$-th round, \texttt{Fed-LWR} calculate layer-wise CKA similarity $\delta_k = \{\delta_k^1, \delta_k^2\ldots,\delta_k^M\}$ between the local model $\boldsymbol{w}_k^t$ and anchor $\boldsymbol{w}_G^t$ averaged by server on hospital $k$. The CKA scores will be used to re-aggregate the local models to obtain fair global model $\boldsymbol{\hat{w}}_G^t$. (b) Variations in CKA similarity versus the layers of local models from two pairs of clients, which are randomly selected. This reveals that the differences between models vary with different layers.}
  \label{fig:fig1} 
\end{figure*}

To address this, some studies~\cite{mohri2019agnostic,hu2022federated,du2021fairness} introduced additional objectives to constrain the optimization process. Similarly, Ditto~\cite{li2021ditto} not only introduced additional objectives but also personalization in the loss function for different clients to improve fairness. In contrast, another promising solution is to re-weight clients by various metrics, such as empirical loss~\cite{li2019fair,li2020tilted} or validation result~\cite{xu2021gradient}, yielding a flexible fairness/accuracy trade-off. In the medical area, FedCE~\cite{jiang2023fair}  also proposed a novel re-weighting strategy by estimating client contribution based on gradients and validation loss. Re-weighting is based on a straightforward idea: assigning higher weights to hospitals with poorer performance, thereby ensuring a more uniform accuracy of the global model~\cite{li2019fair}. Thus, the key is how to dynamically estimate the differences between hospitals. Different from previous re-weighting methods, we try to estimate client weights from a novel perspective based on the characteristics of medical FL. \textbf{\textit{First}}, the private data from different medical institutions is typically collected from various devices, leading to domain shifts across different datasets, resulting in variations in the feature representations of different local models, namely \textit{\textbf{feature shift}}~\cite{li2020fedbn,zhou2022fedfa,huang2023rethinking}. Such variations are the primary factor of the disparate performance of the global model across different datasets. \textbf{\textit{Second}}, previous methods estimated an overall weight for each client, which is a coarse way. As shown in Fig.~\ref{fig:fig1} (b), the degree of differences across local models vary with different layers. Therefore, directly assigning a single weight would ignore these differences in layers and lead to suboptimal aggregation.

In this work, we propose a novel \textbf{L}ayer-\textbf{W}ise \textbf{R}e-weighting method from the perspective of feature shift, namely \texttt{Fed-LWR}, to improve the performance fairness of medical FL. The main idea of \texttt{Fed-LWR} is to quantify the differences in feature representations among different local models. To achieve this, we employ an averaged aggregated global model as the anchor and then estimate the layer-wise centered kernel alignment (CKA)~\cite{cortes2012algorithms,kornblith2019similarity} similarity between the local models and the anchor. 
The estimated similarity reflects the performance differences of the global model across all hospitals. Therefore, we assign larger weights to hospitals with larger feature differences, thereby minimizing the overall performance discrepancies of the global model across different hospitals. The new client weights are used to re-aggregate local models per layer, thereby obtaining a fairer global model. We evaluated our method on two federated medical image segmentation benchmarks. The results indicate that \texttt{Fed-LWR} achieves better fairness/accuracy trade-off compared with several state-of-the-art fair FL methods.

\section{Method}
\subsection{Preliminaries}
\noindent{\textbf{Federated Learning.}} Assume that there are $K$ hospitals participating in a federated learning system, communicating through a trusted central server. Each hospital $k\in[K]$ has a private dataset $D_k$ with $n_k$ training samples $\{\boldsymbol{X}_i, \boldsymbol{Y}_i\}_{i=1}^{n_k}$ and trains a neural network $f = h^M \circ \cdots \circ h^2 \circ h^1$ with $M$ layers. FL typically minimizes empirical risk $\mathcal{L}_k$ to optimize local models and updates the global model through averaging aggregation after each round $t$:
\begin{equation}
    \mathcal{L} = \frac{1}{K}\sum_{k=1}^{K}\mathcal{L}_{k}(\boldsymbol{w}_k^t), \quad \quad \text{and} \quad \boldsymbol{w}_G^t =  \frac{1}{K}  \sum_{k=1}^{K} \boldsymbol{w}_k^t, \label{eq:eq1}
\end{equation}
where $\boldsymbol{w}_G^t$ and $\boldsymbol{w}_k^t$ are the parameters of global and local models. However, in medical FL applications, data from different hospitals are collected using diverse devices, resulting in differences in data distribution.  Such disparities can impair the fairness of the global model across different hospitals. To better understand our
 objective, we introduce the definition of performance fairness from ~\cite{li2019fair}.
\begin{definition}
(Performance Fairness) For two trained models $\boldsymbol{w}$ and $\boldsymbol{\hat{w}}$ from FL, the model $\boldsymbol{\hat{w}}$ provides a more fair solution if the
performance of $\boldsymbol{\hat{w}}$ on the $K$ hospitals is more uniform than the performance of  $\boldsymbol{w}$ on the $K$ hospitals.
\end{definition} 

\noindent{\textbf{Centered Kernel Alignment.}} CKA~\cite{cortes2012algorithms,kornblith2019similarity} is a reliable technique for quantifying the similarity between pairs of neural network representations, which is widely used in various applications~\cite{kim2023stability,raghu2021vision,nguyen2020wide}. Let $z_1$ and $z_2$ be the features from an arbitrary layer of two neural networks,  $\boldsymbol{U}\in \mathbb{R}^{n\times z_1}$ and $\boldsymbol{V}\in \mathbb{R}^{n\times z_2}$ denote their feature  matrices on the same dataset with $n$ samples, $\boldsymbol{K} = \boldsymbol{U}\boldsymbol{U}^\top \in \mathbb{R}^{n\times n}$ and $\boldsymbol{L} = \boldsymbol{V}\boldsymbol{V}^\top \in \mathbb{R}^{n\times n}$  are the Gram matrices of $\boldsymbol{U}$ and $\boldsymbol{V}$. Based on Hilbert-Schmidt Independence
Criterion (HISC)~\cite{gretton2005measuring}, the calculation process of CKA similarity score $\delta \in [0, 1]$  between $\boldsymbol{U}$ and $\boldsymbol{V}$ can be expressed as follows:
\begin{equation}
\delta =\frac{\operatorname{HSIC}(\boldsymbol{K}, \boldsymbol{L})}{\sqrt{\operatorname{HSIC}(\boldsymbol{K}, \boldsymbol{K}) \operatorname{HSIC}(\boldsymbol{L}, \boldsymbol{L})}}, \quad  \operatorname{HSIC}(\boldsymbol{K}, \boldsymbol{L})=\frac{\operatorname{vec}\left(\boldsymbol{K}^{\prime}\right) \cdot \operatorname{vec}\left(\boldsymbol{L}^{\prime}\right)}{(n-1)^2},
\end{equation}
where $\boldsymbol{K}^{\prime}$ and $\boldsymbol{L}^{\prime}$ are centered $\boldsymbol{K}$ and $\boldsymbol{L}$, $\operatorname{vec} (\cdot)$ is vectorization operation. For convenience, we use $\operatorname{CKA}(\cdot)$ to represent the above process.


\subsection{Fed-LWR: Layer-wise Re-weighting for Federated Learning}

 In this section, we propose a novel layer-wise re-weighting method, \texttt{Fed-LWR}, by estimating the differences among clients from the perspective of feature shift. The overview of \texttt{Fed-LWR} is presented in Fig.~\ref{fig:fig1} (a). Compared to standard FL framework~\cite{mcmahan2017communication}, it incorporates additional representation difference estimation on the hospital and layer-wise re-aggregation process on the server after local training. The detailed algorithm is presented in Alg.~\ref{alg:Fed-LWR}.

\noindent{\textbf{Representation Difference Estimation.}} After the local training stage of $t$-th round, we first obtain a global model through Eq.~\eqref{eq:eq1} and send it to each hospital, which serves as an anchor. Then, we compute the CKA similarity between each layer of the local model and the anchor for every client. The above process at client $k$ can be written as: 
\begin{equation}
    \delta_k = [\delta_k^1, \ldots, \delta_k^m, \ldots, \delta_k^M], \quad \quad \text{and} \quad \delta_k^m = \operatorname{CKA}(\boldsymbol{U}_k^{m},\boldsymbol{V}_k^{m}), \label{eq:eq3}
\end{equation}
where $\boldsymbol{U}_k^{m}$ and $\boldsymbol{V}_k^{m}$ are the feature matrices from the $m$-th layer of local model and anchor on dataset $D_k$. 
\begin{equation}
    \begin{split}
    \boldsymbol{U}_k^{m} &=  h^m \circ \cdots \circ h^2 \circ h^1 ([\boldsymbol{X_1}, \boldsymbol{X_2},\ldots, \boldsymbol{X_{n_k}}], \boldsymbol{w}_{k,m}^t\cup\cdots\cup\boldsymbol{w}_{k,2}^t\cup\boldsymbol{w}_{k,1}^t), \\
    \boldsymbol{V}_k^{m} &=  h^m \circ \cdots \circ h^2 \circ h^1 ([\boldsymbol{X_1}, \boldsymbol{X_2},\ldots, \boldsymbol{X_{n_k}}], \boldsymbol{w}_{G,m}^t\cup\cdots\cup\boldsymbol{w}_{G,2}^t\cup\boldsymbol{w}_{G,1}^t),
    \end{split}
\end{equation}
where $\boldsymbol{w}_{k,m}^t$ and $\boldsymbol{w}_{G,m}^t$ are the  $m$-th layer parameters of local model and anchor. CKA can determine correspondences between hidden layers of neural networks trained under different conditions compared to traditional similarity metrics~\cite{kornblith2019similarity}. This best aligns with the situation of FL, as local models are trained on diverse datasets with different distributions.

\noindent{\textbf{Layer-wise Re-aggregation.}} The server collects the CKA similarity scores from each hospital, which reflect the local-global feature differences. 
Lower scores mean a greater difference, indicating that the global model is further from the local optimum, resulting in poorer performance at that hospital. 
Based on this, we convert the CKA similarity scores into aggregation weights:
\begin{equation}
    \rho_k = [\rho^1_k, \ldots, \rho^m_k, \ldots, \rho^M_k], \quad \quad \text{and} \quad \rho^m_k = \frac{1-\delta_k^m}{\sum_{i=1}^{K}(1-\delta_i^m)}.
    \label{eq:eq5}
\end{equation}
Eq.~\eqref{eq:eq5} assigns higher weights to hospitals with greater feature differences, and the sum of the total weight for each layer is 1. Finally, we use the new weights for layer-wise re-aggregation to get the fair global model:
\begin{equation}
    \hat{\boldsymbol{w}}_G^t = \hat{\boldsymbol{w}}_{G,M}^t \cup \cdots \cup \hat{\boldsymbol{w}}_{G,m}^t\cup \cdots \cup \hat{\boldsymbol{w}}_{G,1}^t, \quad \text{and} \quad  \hat{\boldsymbol{w}}_{G,m}^t = \sum_{k=1}^K  \rho^m_k\boldsymbol{w}_{k,m}^t. \label{eq:eq6}
\end{equation}
The fair global model $\hat{\boldsymbol{w}}_G^t$ will serve as the initial weights for the next round of local training.

\begin{algorithm}[t]
\caption{Fed-LWR}
\label{alg:Fed-LWR}
\small
\KwIn{$K$ hospitals, communication rounds $T$, learning rate $\eta$,}
\KwOut{$\hat{\boldsymbol{w}}_G^T$}

Initialize $\boldsymbol{w}^{0}_G$ \quad \tcp{$\hat{\boldsymbol{w}}^{0}_G$ = $\boldsymbol{w}^{0}_G$}
 \For{round $t=1,2,...,T$}{
\For{hospital $k=1,2,...,K$ \textbf{parallelly}}{
     $\boldsymbol{w}^{t}_k$ $\leftarrow$ $\hat{\boldsymbol{w}}^{t-1}_G$ \\
     $\boldsymbol{w}_k^t$ $\leftarrow$  $\boldsymbol{w}^{t}_k - \eta \nabla \mathcal{L}_k$ \quad \tcp{Local training}
}

$\boldsymbol{w}_G^t$ $\leftarrow$ $\frac{1}{K} \sum_{k=1}^{K}$ $\boldsymbol{w}_k^t$ \quad \tcp{Averaging aggregation for the anchor}
\For{hospital $k=1,2,...,K$ \textbf{parallelly}}{
    $\delta_k$ $\leftarrow$ $[\delta_k^1, \ldots, \delta_k^m, \ldots, \delta_k^M]$, \quad $\delta_k^m$ $\leftarrow$ $\operatorname{CKA}(\boldsymbol{U}_k^{m},\boldsymbol{V}_k^{m})$ \quad \tcp{Eq.~\eqref{eq:eq3}}
}
$\rho_k$ $\leftarrow$ $[\rho^1_k, \ldots, \rho^m_k, \ldots, \rho^M_k], \quad \rho^m_k$  $\leftarrow$ $\frac{1-\delta_k^m}{\sum_{i=1}^{K}(1-\delta_i^m)}$ \tcp{Eq.~\eqref{eq:eq6}}

$\hat{\boldsymbol{w}}_G^t$ $\leftarrow$ \{$\hat{\boldsymbol{w}}_{G,m}^t\}_{m=1}^{M}$, $\hat{\boldsymbol{w}}_{G,m}^t$ $\leftarrow$ $\sum_{k=1}^K \rho^m_k\boldsymbol{w}_{k,m}^t$ \tcp{Eq.~\eqref{eq:eq3}}
}
return $\hat{\boldsymbol{w}}_G^T$
\end{algorithm}

\section{Experiment}
\subsection{Experimental Setup}
\noindent{\textbf{Datasets.}} To evaluate the effectiveness of our method, we conducted experiments on two medical image segmentation datasets: \textbf{ProstateMRI}~\cite{liu2020ms} and \textbf{RIF}~\cite{wang2020dofe}, which are widely used in medical FL~\cite{jiang2023fair,jiang2023iop,wang2023feddp,wang2022personalizing,zhou2022fedfa}. ProstateMRI collects T2-weighted MRI images from six different data sources for prostate segmentation. The size of all images has been processed to 384 $\times$ 384. We treat each data source as a client and divide the data into training, validation, and testing sets with a ratio of 6:2:2~\cite{zhou2022fedfa}. RIF contains the retinal fundus images from four different clinical institutions for optic disc and cup segmentation. Following ~\cite{wang2023feddp,wang2022personalizing}, we resize all images to 384 $\times$ 384. Since each sub-dataset has already been pre-divided into training and testing sets~\cite{wang2020dofe}, we further split the training set into training and validation sets at a ratio of 4:1.

\noindent{\textbf{Baselines.}} We compared our method with following baselines: \ding{182} \textbf{Solo}: clients train models locally without communication;
\ding{183} \textbf{FedAvg}~\cite{mcmahan2017communication}: the most popular FL method which updates the global model through parameter averaging; and several state-of-the-art fair FL methods including \ding{184} \textbf{q-FedAvg}~\cite{li2019fair}: a method re-weights the clients through empirical loss; \ding{185} \textbf{CFFL}~\cite{lyu2020collaborative}:  it re-allocates the received model from the server by estimating the contribution of clients from validation results; \ding{186} \textbf{CGSV}~\cite{xu2021gradient}: it proposed a cosine gradient Shapley value to estimate the contribution of clients from the gradient; \ding{187} \textbf{Ditto}~\cite{li2021ditto}: a personalized fair FL method to set different optimization objectives for clients; \ding{188} \textbf{FedCE}~\cite{jiang2023fair}: a fair federated medical image segmentation method by estimating the weights of clients simultaneously from gradients and validation loss. 
There are two different versions, \textit{i.e.}, FedCE (Sum.) and FedCE (Multi.), that use addition and multiplication to merge contributions from gradients and losses, respectively. We use the Dice coefficient, a popular metric in medical image segmentation tasks, to evaluate the performance of the method. Following ~\cite{li2019fair,li2021ditto}, we evaluate the performance fairness of methods by the standard deviation of testing performance across all clients.

\noindent{\textbf{Implementation Details.}} We implement all methods using PyTorch and conduct all experiments using an NVIDIA RTX 4090 GPU with 24GB of memory. Besides, we use U-Net~\cite{ronneberger2015u} for the segmentation task and Dice loss as the optimization objective for the clients. The network is optimized by the Adam optimizer with a learning rate of 1e-3 and weight decay of 1e-4. The batch size is set to 8. We run 200 communication rounds with 1 local epoch and ensure all methods have converged stably. For a fair comparison, all methods adopt the same experimental settings.

\subsection{Comparison with State-of-the-Arts}

In Tables~\ref{tab:table1} and ~\ref{tab:table2}, we present the results of quantitative comparison for all methods on ProstateMRI and RIF, including the Dice scores on the testing sets of each client, as well as their average results and standard deviation. Apparently, with the proposed layer-wise re-weighting strategy, \texttt{Fed-LWR} significantly improves the performance of FedAvg from 88.78\% to \textbf{93.25\%} on ProstateMRI and from 84.63\% to \textbf{87.17\%} on RIF. Meanwhile, it also outperforms other fair FL methods on both two datasets. The above results indicate that our method can effectively alleviate the feature shift problem. Notably, FedAvg achieved lower performance on RIF compared to Solo, which is due to the high heterogeneity of client data. In terms of performance fairness, all fair FL methods show better fairness compared with FedAvg. However, the fairness of some methods (\textit{e.g.}, CFFL and CGSV) is achieved by sacrificing global performance, indicating that their fairness mechanisms may hinder the convergence of the global model to some extent. Compared to other methods, \texttt{Fed-LWR} achieved the best performance fairness on two datasets. This is attributed to \texttt{Fed-LWR} improving the performance of worse clients, \textit{e.g.}, it improves the performance of \texttt{Site1} from 82.70\% to \textbf{85.89\%} and \texttt{Site2} from 72.68\% to \textbf{79.43\%} on RIF. The above results demonstrate the effectiveness of \texttt{Fed-LWR} in improving the fairness performance of the FL system. 

\begin{table*}[!t]
\centering
\caption{\textbf{Quantitative comparison}  with Dice coefficient (\%) on the testing set of ProstateMRI~\cite{liu2020ms}.
We report the performance of six clients, as well as their average result (\texttt{Avg.}) and standard deviation (\texttt{Std.}). The best results are marked in bold. }
\setlength\tabcolsep{5pt}
\renewcommand{\aboverulesep}{0pt}
\renewcommand{\belowrulesep}{0pt}
\setlength\cellspacetoplimit{5pt}
\setlength\cellspacebottomlimit{5pt}
\renewcommand\arraystretch{1.2}{
\begin{tabular}{l |cccccc|cc}
\toprule
\rowcolor[HTML]{EFEFEF} \textbf{Method}  & \textbf{\texttt{Site1}}  &  \textbf{\texttt{Site2}}  &  \textbf{\texttt{Site3}}  &  \textbf{\texttt{Site4}}  & \textbf{\texttt{Site5}}  &  \textbf{\texttt{Site6}}  & \textbf{\texttt{Avg.}} & \textbf{\texttt{Std.}} \\
\hline
Solo & 84.41 & 85.02 & \textbf{93.70} & 91.06 & 89.33 & 85.43 & 88.16 & 3.46 \\

\hline

FedAvg~\cite{mcmahan2017communication}  & 85.84 & 92.27 & 93.58 & 87.89 & 92.47 & 80.66 & 88.78 & 4.54 \\

q-FedAvg~\cite{li2019fair} & 85.84 & \textbf{94.93} & 90.22 & 88.53 & 90.43 & 83.32 & 88.88 & 3.67 \\
CFFL~\cite{lyu2020collaborative} & 82.84 & 94.49 & 89.83 & 86.89 & 90.83 & 82.59 & 87.91 & 4.29 \\
CGSV~\cite{xu2021gradient} & 82.36 & 91.02 & 90.14 & 87.25 & 91.06 & 82.71 & 87.42 & 3.68 \\
Ditto~\cite{li2021ditto} & 88.15 & 93.49  & 92.72  & 89.99  & 92.62 & 83.64  & 90.10  & 3.42 \\
FedCE (Sum.) ~\cite{jiang2023fair} & 88.13 & 94.02 &	91.14 &	90.63 &	92.27 &	87.93 &	90.69 &	2.15 \\
FedCE (Multi.) ~\cite{jiang2023fair} & 89.85 & 93.49 &	92.21 &	90.67 &	93.66 & 89.53 &	91.57 &	1.65 \\
\hline
\texttt{Fed-LWR} & \textbf{91.92} &	94.73 & 93.22 &	\textbf{93.34} & \textbf{94.06} & \textbf{92.21} & \textbf{93.25} &	\textbf{0.97} \\

\bottomrule
\end{tabular}}
\label{tab:table1}
\end{table*}

\begin{table*}[!t]
\centering
\caption{\textbf{Quantitative comparison}  with Dice coefficient (\%) on the testing set of RIF~\cite{wang2020dofe}.
We report the performance of four clients, as well as their average result (\texttt{Avg.}) and standard deviation (\texttt{Std.}). The best results are marked in bold.}
\setlength\tabcolsep{5pt}
\renewcommand{\aboverulesep}{0pt}
\renewcommand{\belowrulesep}{0pt}
\setlength\cellspacetoplimit{5pt}
\setlength\cellspacebottomlimit{5pt}
\renewcommand\arraystretch{1.2}{
\begin{tabular}{l|cccc|cc}
\toprule
\rowcolor[HTML]{EFEFEF} \textbf{Method}  & \textbf{\texttt{Site1}}  &  \textbf{\texttt{Site2}}  &  \textbf{\texttt{Site3}}  &  \textbf{\texttt{Site4}}   & \textbf{\texttt{Avg.}} & \textbf{\texttt{Std.}} \\
\hline
Solo & 82.01 & 74.57 & 91.26 & \textbf{92.56}& 85.10 & 7.31\\

\hline

FedAvg~\cite{mcmahan2017communication}  & 82.70 & 72.68 & 91.19 & 91.93 & 84.63 & 7.79\\

q-FedAvg~\cite{li2019fair}  & 77.83 & \textbf{80.48} & 91.72 & 91.62 & 85.41 & 6.32 \\
CFFL~\cite{lyu2020collaborative} & 81.49 & 74.97 & 91.33 & 89.53 & 84.33 & 6.55\\
CGSV~\cite{xu2021gradient}  & 78.79 & 77.30 & 89.77 & 91.78 & 84.41 & 6.42\\
Ditto~\cite{li2021ditto} & 82.11 & 78.91 & 90.90 & 91.95 & 85.97 & 5.58 \\
FedCE (Sum.)~\cite{jiang2023fair}   & \textbf{87.02} & 76.60 & 90.92 & 90.28 & 86.21 & 5.73 \\
FedCE (Multi.)~\cite{jiang2023fair} & 86.00 & 77.42 & 90.69 & 89.93 & 86.01 & 5.26\\
\hline
\texttt{Fed-LWR} & 85.89 & 79.43 & \textbf{92.42} & 90.96 & \textbf{87.17} & \textbf{5.08}\\

\bottomrule
\end{tabular}}
\label{tab:table2}
\end{table*}

\subsection{Analytical Studies}
\noindent{\textbf{Ablation Study.}} To provide more insights into \texttt{Fed-LWR}, we further delve into the design of our approach, \textit{i.e.}, measurement of feature difference and layer-wise aggregation. First, we build up two variants of our method as follows: \ding{182} \texttt{Fed-LWR-v1}: we utilize cosine similarity, widely used in various applications~\cite{radford2021learning,xie2022c2am}, instead of CKA similarity to measure feature difference for \texttt{Fed-LWR};  \ding{183} \texttt{Fed-LWR-v2}: we only use the CKA similarity of the last layer of the U-Net encoder to estimate the weights of clients and aggregate the entire model. As shown in Table~\ref{tab:table3}, the performance significance of cosine similarity is significantly lower than that of CKA similarity because cosine similarity is more suitable as an optimization objective to minimize the feature distance, and is difficult to estimate the similarity of two unrelated features. Besides, the comparison results between \texttt{Fed-LWR} and \texttt{Fed-LWR-v2} demonstrate the importance of layer-wise aggregation.

\begin{table*}[!t]
\centering
\caption{\textbf{Results of ablation study } on the testing set of ProstateMRI~\cite{liu2020ms}.}
\setlength\tabcolsep{5pt}
\renewcommand{\aboverulesep}{0pt}
\renewcommand{\belowrulesep}{0pt}
\setlength\cellspacetoplimit{5pt}
\setlength\cellspacebottomlimit{5pt}
\renewcommand\arraystretch{1.2}{
\begin{tabular}{l |cccccc|cc}
\toprule
\rowcolor[HTML]{EFEFEF} \textbf{Method}  & \textbf{\texttt{Site1}}  &  \textbf{\texttt{Site2}}  &  \textbf{\texttt{Site3}}  &  \textbf{\texttt{Site4}}  & \textbf{\texttt{Site5}}  &  \textbf{\texttt{Site6}}  & \textbf{\texttt{Avg.}} & \textbf{\texttt{Std.}} \\
\hline
FedAvg~\cite{mcmahan2017communication}  & 85.84 & 92.27 & 93.58 & 87.89 & 92.47 & 80.66 & 88.78 & 4.54 \\
\texttt{Fed-LWR-v1} & 85.76& 94.29 & \textbf{94.36} & 88.51 & 92.91 & 86.33 & 90.36 & 3.62\\
\texttt{Fed-LWR-v2} & 89.27& \textbf{94.86} & 92.99 & 90.50 & 93.24 & 89.32& 91.70 & 2.12 \\
\hline
\texttt{Fed-LWR} & \textbf{91.92} &	94.73 & 93.22 &	\textbf{93.34} & \textbf{94.06} & \textbf{92.21} & \textbf{93.25} &	\textbf{0.97} \\

\bottomrule
\end{tabular}}
\label{tab:table3}
\end{table*}

\begin{figure*}[t]
\centering
  \includegraphics[width=0.85\textwidth]{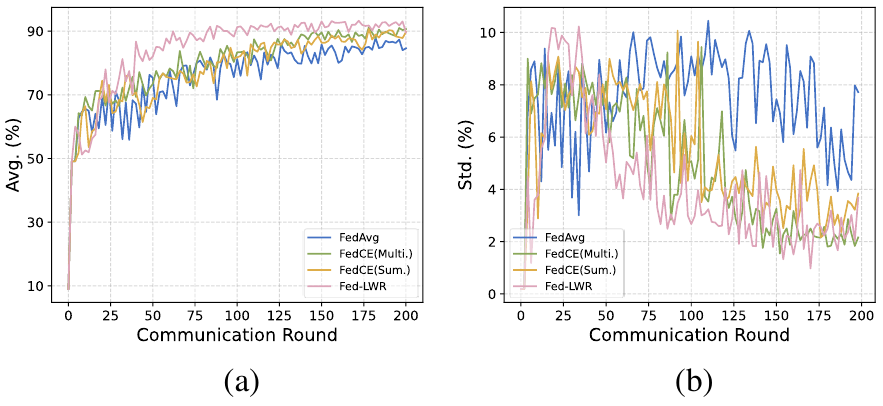} \caption{\textbf{(a) Avg. and (b) Std. } versus the number of communication rounds on the testing set of ProstateMRI~\cite{liu2020ms}.}
  \label{fig:fig2} 
\end{figure*}

\noindent{\textbf{Convergence.}} We visualize the curve of testing performance versus communication rounds for FedAvg, FedCE, and our method. As shown in Fig.~\ref{fig:fig2} (a), we can observe that \texttt{Fed-LWR} converges faster  (stable after 90 rounds) compared to the other two methods. Interestingly, from Fig.~\ref{fig:fig2} (b), \texttt{Fed-LWR} exhibits a larger standard deviation between rounds 20 to 50, as it is converging rapidly. This indicates that the fairness mechanism of \texttt{Fed-LWR} does not slow its convergence.

\section{Conclusion}
In this work, we improved the performance fairness of medical FL from the perspective of feature shift. Specifically, we proposed a novel fair FL framework, \texttt{Fed-LWR}, by estimating the feature differences between local models and the global model for layer-wise re-aggregation. 

\noindent{\textbf{Future Work.}} We primarily focused our evaluation on segmentation tasks, as it is the most representative task in medical image analysis. Nevertheless, \texttt{Fed-LWR} is a versatile framework, and we will further evaluate its effectiveness on more medical image tasks (\emph{e.g.}, classification or reconstruction) in future work.

\subsubsection{Acknowledgments.} 
This work is supported by the Guangzhou-HKUST(GZ) Joint Funding Program (No. 2023A03J0671), the Guangzhou Municipal Science and Technology Project (Grant No. 2023A03J0671), National Research Foundation, Singapore under its AI Singapore Programme (AISG Award No: AISG2-TC-2021-003), and the Agency for Science, Technology and Research (A*STAR) through its AME Programmatic Funding Scheme Under Project A20H4b0141. 

\subsubsection{Disclosure of Interests.} The authors have no competing interests to declare that are relevant to the content of this article.

%
%
%
\bibliographystyle{splncs04}
\bibliography{mybib}
\end{document}